# Robust Path Following on Rivers Using Bootstrapped Reinforcement Learning

Niklas Paulig[a,*], Ostap Okhrin[a]

[a]*Technische Universität Dresden, Chair of Econometrics and Statistics, esp. in the Transport Sector, Dresden, 01187, Saxony, Germany*

**ABSTRACT**

This paper develops a Deep Reinforcement Learning (DRL)-agent for navigation and control of autonomous surface vessels (ASV) on inland waterways. Spatial restrictions due to waterway geometry and the resulting challenges, such as high flow velocities or shallow banks, require controlled and precise movement of the ASV. A state-of-the-art bootstrapped Q-learning algorithm in combination with a versatile training environment generator leads to a robust and accurate rudder controller. To validate our results, we compare the path-following capabilities of the proposed approach to a vessel-specific PID controller on real-world river data from the lower- and middle Rhine, indicating that the DRL algorithm could effectively prove generalizability even in never-seen scenarios while simultaneously attaining high navigational accuracy.

## 1. Introduction

According to a survey on the development of the global ocean surface robot market by BIS Research (2018), the market for autonomous vessels is "expected to grow at the rate of 26.7% for the period 2024-2035". For autonomous vessels to be integrated seamlessly into existing hybrid traffic, it is crucial to fulfilling their automated tasks with high accuracy. This is especially true for spatially restricted inland waterways such as rivers, bights and canals. The Directorate-General for Mobility and Transport (2023) of the European Commission emphasizes the importance of inland waterway traffic and its development due to decreased costs and increased safety in comparison to other modes of transport. To build on this directive, the present study is one of the first approaches to solving the path-following problem for underactuated vessels on restricted waterways using deep reinforcement learning (DRL) and under consideration of environmental influences. Breivik and Fossen (2004) stated, that, compared to other automated systems, ships on inland waterways face additional challenges due to their environment (e.g., strong directional currents, shallow banks) and underlying physics (e.g., underactuation, highly non-linear maneuvering models), leading to a highly dynamic and stochastic operational environment. To overcome these hurdles, we incorporate water depth, current direction, and speed into the agent's perception, allowing it to navigate tight river turns safely.

In this paper, an ensemble-based DRL algorithm is used to develop a high-precision and generalizable path-following controller for inland transportation vessels. The contribution to the field is as follows:

- We develop a tunable segmental generator to create realistic and diverse training environments specifically for inland waterways. The source code is publicly available as a GitHub repository via `github.com/nikpau/sr-gen`.

- We use a state-of-the-art bootstrapped DQN-based algorithm to generate robust and generalizable policies for rudder control under varying external environmental disturbances.

- To demonstrate the generalizability and robustness of our approach, we validate the produced policies on real-world data from the middle and lower Rhine.

The rest of the paper is organized as follows. Section 2 recapitulates current literature on the topic of path-following and formalizes the problem. The kinodynamic ship-maneuvering model is detailed in Section 3 while section 4 introduces methodologies and how they are incorporated into the path-follower controller design. The fifth section sets up a benchmark controller for validation and Section 6 applies the path-following results to various maritime scenarios

---

*Corresponding author

niklas.paulig@tu-dresden.de (N. Paulig); ostap.okhrin@tu-dresden.de (O. Okhrin)
ORCID(s): 0000-0002-0220-6702 (N. Paulig)
https://www.linkedin.com/profile/view?id=niklaspaulig0 (N. Paulig)

and validates the controller on separate segments of the Rhine river. Section 7 performs a robustness analysis, and the last section concludes.

## 2. Path-following for ASV

### 2.1. State of the art

The objective of path-following for ships demands a controller to generate steering commands that enable an underactuated autonomous surface vessel (ASV) to follow a pre-defined path with minimal angular and spatial deviation. The problem formulation for this study will only include rudder angles as control outputs while keeping the engine revolutions constant.

According to Fossen (2011), an onboard path-following system requires three sub-systems to be implemented: *guidance*, *navigation*, and *control*. To autonomously control a vessel, we require to know its current position (navigation), planned trajectory (guidance), and a set of control actions to move towards its current goal (control).

Line-of-sight (LOS) guidance is one common approach in implementing directional awareness of the agent, achieving convergence to the desired path. It has been successfully applied in various problem settings, as in Fossen et al. (2003) and Fossen and Lekkas (2017) with traditional control approaches and Oh and Sun (2010) for a model-predictive-control application. Vector field guidance (VFG) Nelson et al. (2007) is a different approach that uses a global vector field encompassing the path to guide the vessel towards it, independent of the magnitude of deviation. Woo et al. (2019) integrated VFG into a combined path-following and collision avoidance method.

After setting up a suitable guidance algorithm, the next step demands a control system. There are two main methodological approaches to solving the path-following problem, analytic control, and reinforcement learning. As part of the analytic control family, proportional-integral-derivative (PID) controllers are well understood, require few computational resources and have successfully been used to develop path-followers in calm and disturbed waters. While Moreira et al. (2007) achieved path-following using a LOS guidance system and PID controller for steering control, Perera et al. (2014) used fuzzy logic to derive, and PID controller to execute sequential actions for path-following but also collision avoidance of a small model vessel. Paramesh and Rajendran (2021) used PID control to navigate a tanker along a given path under the influence of regular waves. PID performance, however, is vessel-dependent, requires expert tuning and often is sensitive to external disturbances. More recent advances in control theory allow for different approaches such as non-linear model-predictive-control Xia et al. (2013); Sandeepkumar et al. (2022), backstepping control Zhang et al. (2017), or sliding mode control Liu et al. (2018).

Reinforcement Learning (RL) is based on agent-environment interaction, aimed at learning a correct set of actions given some observed state. The actions taken by the agent are evaluated based on a hand-crafted reward function, whose goal is to reinforce actions that bring the agent closer to its defined goal, ultimately finding a policy that solves the problem optimally. Recently, interest in academia in using RL-based motion control has surged due to its ability to tackle problems with high uncertainty and non-linear system dynamics. Various researchers, such as Shen and Guo (2016), Martinsen and Lekkas (2018a) use a family of continuous-action algorithms in which the agent is free to choose any action on the applicable range every time step. While this allows for a highly reactive policy, it is possible to choose actions leading to unrealistic behavior, such as maximally opposing rudder angles on two successive time steps. Discrete action solutions as discussed by Zhao et al. (2019); Amendola et al. (2019, 2020); Martinsen et al. (2020) often share the identical drawback, leading some researchers to block the agent from choosing the next action until the last one was performed subject to physical constraints of the vessel. While this approach is viable in restricting physically impossible movements, it impairs the agent's reactivity during the time of blockage. To mitigate this problem in this study, we opted only to allow the agent to choose from actions within the vessel's physical possibilities.

### 2.2. Path following on rivers

Contrary to the open sea, rivers pose additional navigational challenges mainly due to their limited spatial extent, strong directional currents, shallow banks, and small path-curve radii.

Figure 1 depicts the heading-control setup used in this study. We assume a given path consisting of a discrete set of $K$ waypoints $P_k = (x_k, y_k)^\top, k \in \{1, \ldots, K\}$, where two consecutive waypoints $P_k$ and $P_{k+1}$ enclose the path heading

$$\chi_{P_k} = \operatorname{atan2}(y_{k+1} - y_k, x_{k+1} - x_k), \tag{1}$$

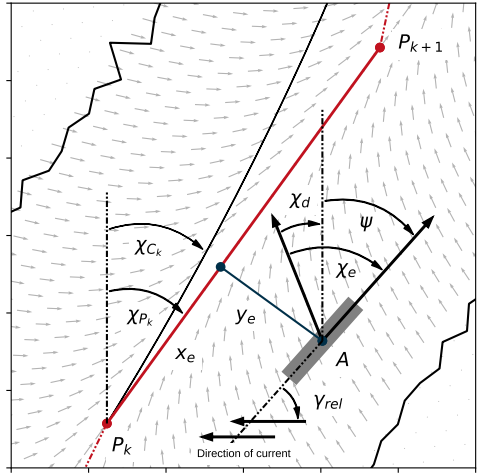

**Figure 1:** Heading control setup

Using such discrete waypoints, however, will produce discontinuous jumps in the desired path heading once the vessel crosses the waypoint in front of it. To infer a continuous path heading during training, this study uses a distance-dependent weighted sum of the current and next path segment heading:

$$\chi_{C_k} = \left\{ 1 - \frac{x_e}{d(P_k, P_{k+1})} \chi_{P_k} \right\} + \left\{ \frac{x_e}{d(P_k, P_{k+1})} \chi_{P_{k+1}} \right\}, \tag{2}$$

with $d(P_k, P_{k+1}) = \sqrt{(x_k - x_{k+1})^2 + (y_k - y_{k+1})^2}$ being the Euclidean distance between two succeeding waypoints and the along-track distance $x_e$ given by

$$x_e = (x_A - x_k) \cos\left(\chi_{P_k}\right) + (y_A - y_k) \sin\left(\chi_{P_k}\right), \tag{3}$$

using the current vessel position $A = (x_A, y_A)^\top$. Most vessel-related variables such as the vessel position, the along-track error, cross-track error etc., are time-dependent. For simplicity, and to avoid clutter, we will drop the time index $t$ in this and the next section, i.e. we write $A = (x_A, y_A)^\top$ instead of $A_t = (x_{A,t}, y_{A,t})^\top$.

There are two fundamental metrics to control for in a path-following scenario: *Cross-track-error* ($y_e$) and *heading-error* ($\chi_e$). The cross-track-error normal to the path can then be found via

$$y_e = (x_A - x_k) \sin(\chi_{C_k}) + (y_A - y_k) \cos(\chi_{C_k}). \tag{4}$$

From the cross-track-error, we can construct a vector field after Nelson et al. (2007) to determine the desired course as

$$\chi_d = \tan^{-1}(c y_e) + \chi_{P_k}, \tag{5}$$

where $c$ is a tunable hyperparameter controlling the speed of convergence of the vector field. Using the vessel's current heading $\psi$ and drift angle $\beta$ (see Figure 2), the course error calculates to

$$\chi_e = \chi_d - \psi - \beta. \tag{6}$$

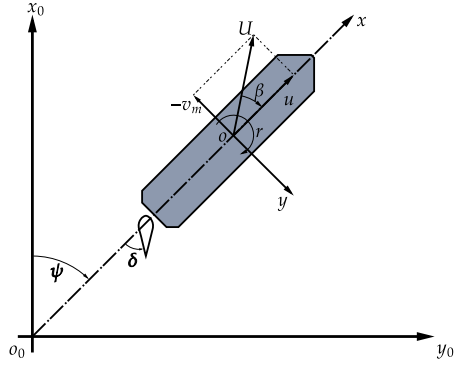

**Figure 2:** Global and local coordinate systems.

The path-following objective is to generate control actions that move both cross-track error and course error to zero.

## 3. ASV kinodynamics
### 3.1. Equations of motion

The present study uses the 3-degree-of-freedom MMG model of ship maneuvering Yasukawa and Yoshimura (2015) to describe the autonomous vessel's movement in the horizontal plane. The ASV is modeled as a rigid body with a single propeller. This paper uses the coordinate system shown in Figure 2. The $o_0 - x_0 y_0 z_0$ coordinate system corresponds to the earth-fixed water surface while the $o - xyz$ system is vessel-fixed with origin $o$ at midship and $x, y$ pointing towards the bow and starboard respectively, $z$ is pointing downwards. The center of gravity is at $(x_G, 0, 0)$ in the vessel-fixed coordinate system; total sway at the center of gravity then is $v = v_m + x_G r$, with $v_m$ being the sway velocity at midships and $r$ the turning rate. Surge velocity is denoted by $u$, thus total ship velocity is given by $U = \sqrt{u^2 + v_m^2}$, drift angle at midships by $\beta = \tan^{-1}(v_m/u)$ and the heading $\psi$ by the angle between $x_0$ and $x$.

The forces acting on the ship are decomposed as follows:

$$\left.\begin{aligned}
(m + m_x)\dot{u} - (m + m_y) v_m r - x_G m r^2 &= X, \\
(m + m_y)\dot{v}_m + (m + m_x) ur + x_G m \dot{r} &= Y, \\
(I_{zG} + x_G^2 m + J_z)\dot{r} + x_G m (\dot{v}_m + ur) &= N,
\end{aligned}\right\} \quad (7)$$

where $m$ is the mass of the ASV, $m_x$ and $m_y$ are the added masses in $x$- and $y$-axis direction respectively, $x_G$ is the longitudinal coordinate of center of gravity, $I_{zG}$ is the moment of inertia, $J_z$ is the added moment of inertia, and $r$ is the yaw rate.

Total forces of the left-hand-side of (7), $X, Y, N$, are surge force, lateral force and yaw moment around midship and consist of the following parts:

$$\left.\begin{aligned}
X &= X_H + X_R + X_P, \\
Y &= Y_H + Y_R, \\
N &= N_H + N_R.
\end{aligned}\right\} \quad (8)$$

The subscripts H, R, P describe forces acting on the hull, rudder and propeller respectively. Further implementation details are deferred to the original paper by Yasukawa and Yoshimura (2015).

| | |
|---|---|
| Scale | 1/5 |
| Displacement | 2500.8 $m^3$ |
| Length between perpendiculars | 64.0 $m$ |
| Width | 11.6 $m$ |
| Block coefficient | 0.81 |
| Draft | 4.16 $m$ |
| Rudder area | 4.5 $m^2$ |
| Propeller diameter | 1.76 $m$ |

**Table 1**
Principal particulars of a KVLCC2 L64-model tanker

## 3.2. Environmental forces

According to Fossen (2011, p. 39), vessel speed under the influence of currents becomes a relative speed $U = \sqrt{(u - u_c)^2 + (v_m - v_c)^2}$ where $u_c$ and $v_c$ are the current component velocities in longitudinal and lateral direction.

The effects of shallow water on wake fraction, thrust deduction and flow-straightening coefficients are calculated after Amin and Hasegawa (2010) while the effects on hydrodynamic derivatives are adapted using combined formulations from Kijima and Nakiri (1990) and Ankudinov et al. (1990). A summary can be found at Taimuri et al. (2020). The effects on the maneuverability of the vessel are demonstrated in a zigzag and turning maneuver test shown in Figure 3. For the zigzag test, the vessel starts with an initial velocity $U_0 = 4.0 m/s$, a rudder angle of 0° and an arbitrary course $\bar{\psi} - \beta$ (This study uses $\bar{\psi} - \beta = 0$). The rudder angle is increased by $5.0° s^{-1}$ until it reaches its maximum value (10° or 20°), at which it is held until the vessel's course is changed by the same amount. The rudder direction is then reversed with the same principle. For this test, currents are turned off. The turning maneuver test starts with the same initial conditions as the zigzag test, however, the rudder angle is increased to 35° and held there for the rest of the experiment. For both tests, we see impaired maneuverability for the vessel under shallow water conditions ($h/d = 1.2$), which is to be expected and emphasizes the need for a precise controller on inland waterways.

The open-source implementation of the MMG dynamics used for this study can be found at Paulig (2022).

## 3.3. Vessel model

The vessel type used for simulation is a 1:5 scale model (L-64) of the KVLCC2 Tanker, as it has one of the most well-understood dynamics publicly available. The ship's principal particulars can be found in Table 1. We use a 1:5 scaling to mimic the dimensions and behavior of small- to medium-sized inland cargo vessels.

## 4. Reinforcement Learning framework

### 4.1. Fundamentals

RL is a subfield of machine learning in which an agent is trained to act in an environment such that it maximizes a reward signal received from the environment Sutton and Barto (2018). Formally, the simulated environment is modeled as a Markov Decision Process (MDP) Puterman (1994) described by the tuple $(\mathcal{S}, \mathcal{A}, \mathcal{P}, \mathcal{R}, \gamma)$. At every time step $t$, given a current state of the environment $s_t \in \mathcal{S}$ the agent executes an action $a_t \in \mathcal{A}$ according to a parameterized policy $\pi_\theta : \mathcal{S} \to \mathcal{A}$. After performing the action, the agent receives a reward $r_t \in \mathcal{R}$ and transitions to the next state $s_{t+1}$ according to the state transition probability distribution $\mathcal{P} : \mathcal{S} \times \mathcal{A} \times \mathcal{S} \to [0, 1]$. The return is defined as the cumulative discounted reward $R_t = \sum_{i=t}^{T} \gamma^{i-t} r_i$ from the current time step until the final time step of the episode $t + T$ with $\gamma \in [0, 1]$ being the discount factor that trades off the importance of immediate and later rewards. All the contributions of the tuple $(\mathcal{S}, \mathcal{A}, \mathcal{P}, \mathcal{R}, \gamma)$ for the path-following objective will be specified in Section 4.3.

The goal of RL is to find a policy that maximizes reward in the long-term starting from some initial state. Most current algorithms use a state-action value function $Q : \mathcal{S} \times \mathcal{A} \to \mathbb{R}$ to assign a value to each state-action pair such that higher values represent pairs leading to a higher long-term reward. $Q^\pi(s, a) = \mathbb{E}_{s \sim \mathcal{P}, a \sim \pi}(R_t \mid s_t = s, a_t = a)$ resembles the expected discounted sum of rewards starting from state $s$, taking action $a$ and following policy $\pi$ afterwards. The algorithmic foundation for this work is the Q-Learning algorithm Watkins and Dayan (1992) that uses the Bellman

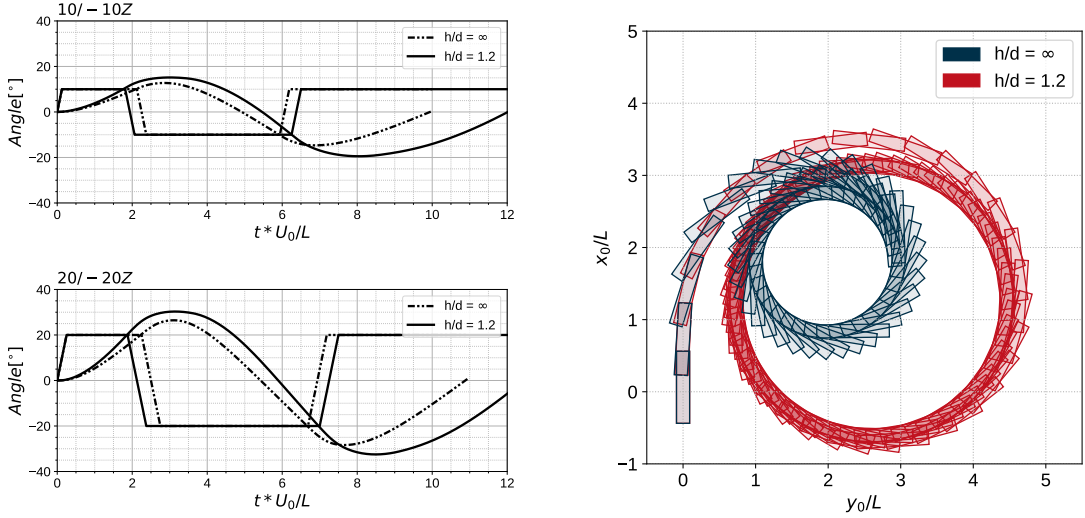

(a) Zigzag maneuvers for 10 and 20-degree rudder angles in shallow and deep waters. Initial velocity $U_0 = 4.0 m/s$, steering rate $\Delta \delta = 5.0°/s$.

(b) 35° starboard turning maneuver in deep and shallow water. Initial velocity $U_0 = 4.0 m/s$, steering rate $\Delta \delta = 5.0°/s$.

**Figure 3:** Zigzag and turning maneuver tests for water depth $h$ and ship draught $d$.

optimality equations Bellman (1957) to solve for the optimal Q-values $Q^*$ satisfying

$$Q^*(s,a) = \mathbb{E}\left\{ r_t + \max_{a_{t+1} \in \mathcal{A}} Q^*\left(s_{t+1}, a_{t+1}\right) \mid s_t = s, a_t = a \right\}, \tag{9}$$

from which an optimal policy can be derived by $\pi^*(s) = \operatorname{argmax}_a Q^*(s,a)$. The Q-Learning update rule for a given Q-value estimate $\hat{Q}(s,a)$ and learning rate $\alpha$ is

$$\hat{Q}\left(s_t, a_t\right) \leftarrow \hat{Q}\left(s_t, a_t\right) + \alpha\{\bar{\tau}^T - \hat{Q}\left(s_t, a_t\right)\}, \text{ for } \bar{\tau}^T = r_{t+1} + \gamma \max_{a_{t+1} \in \mathcal{A}} \hat{Q}\left(s_{t+1}, a_{t+1}\right). \tag{10}$$

To keep track of the Q-values, their value must be stored for each state-action pair, which is infeasible even for moderately-sized environments. To overcome this limitation Mnih et al. (2015) introduced the DQN algorithm that uses two neural networks as function approximators for Q-Value estimation. The second network is a frozen copy of the first one that gets periodically updated to match the first. This contributes significantly to training stability, as bootstrapping the next action's Q-Value from the same network can lead to unpredictable behavior, especially in the early stages of training.

### 4.2. Overestimation bias

One of the most often criticized problems of the Q-learning update rule in (10), is an overestimation bias. It is induced through the fact, that the estimation of the bootstrapped target $\bar{\tau}^T$ uses the maximum over all possible actions. Because all Q-Values are approximations of their true expectation, some estimations are probably higher than the true expected value Thrun and Schwartz (1993). This can lead to misjudgment during exploration, as states with falsely attributed high Q-Values are taken into consideration more often, than the ones with falsely attributed low Q-Values. Eventually, this inequality can overall lead to suboptimal policies.

Several approaches set out to mitigate this overestimation. Van Hasselt (2010) introduced Double Q-Learning, replacing over- with underestimation by separating selection and evaluation of the maximum, thereby achieving significant performance improvements in the deep-learning setup Van Hasselt et al. (2016).

The approach in this paper was proposed by Waltz and Okhrin (2022) and provides an extension of the bootstrapping framework from Osband et al. (2016). The general idea is to rely on the ability of bootstrapping to provide measures of accuracy for statistical estimates, which is usually achieved by resampling an original dataset with replacement and calculating the statistics of interest using these bootstrapped samples.

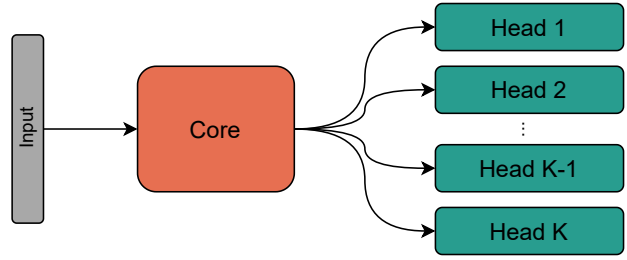

**Figure 4:** Bootstrapped DQN architecture as proposed by Osband et al. (2016)

In the DRL setting, this is translated into either maintaining several distinct Q-networks, each with its own target network or using a single network with a shared core and several heads (see Figure 4). This work will use the latter approach. The bootstrapping nature is achieved by randomly initializing the network heads and using a binary map to determine which head is to be updated on the current iteration. Additionally, Waltz and Okhrin (2022) propose to replace the maximum over all possible actions in the target with a kernel-based testing procedure. Suppose a network with one common core and $B \in \mathbb{N}$ heads. Furthermore, let $\kappa$ be a kernel function (in our study we use the Gaussian cumulative distribution function $\Phi(\cdot)$). The target for the $b^{th}$ head then becomes

$$\bar{\tau}^{T,b} = r + \gamma \left[ \sum_{a_{t+1} \in \mathcal{A}} \kappa \left\{ T_{\hat{Q}_b} \left( s_{t+1}, a_{t+1} \right) \right\} \right]^{-1} \sum_{a_{t+1} \in \mathcal{A}} \kappa \left\{ T_{\hat{Q}_b} \left( s_{t+1}, a_{t+1} \right) \right\} \hat{Q}_b \left( s_{t+1}, a_{t+1}; \theta_b^- \right), \quad (11)$$

where

$$T_{\hat{Q}_b} \left( s_{t+1}, a_{t+1} \right) = \frac{\hat{Q}_b \left( s_{t+1}, a_{t+1}; \theta_b^- \right) - \max_{a_{t+1} \in \mathcal{A}} \hat{Q}_b \left( s_{t+1}, a_{t+1}; \theta_b^- \right)}{\sqrt{\widehat{\text{Var}} \left\{ \hat{Q}_b \left( s_{t+1}, a_{t+1}; \theta_b^- \right) \right\} + \widehat{\text{Var}} \left\{ \hat{Q}_b \left( s_{t+1}, a^*; \theta_b^- \right) \right\}}}, \quad (12)$$

is a statistic for testing whether the selected action from head $v$ is not smaller than the maximum estimate for that head, and the currently maximizing action

$$a^* \in \left\{ a \in \mathcal{A} \mid \hat{Q}_b \left( s_{t+1}, a; \theta_b^- \right) = \max_{a_{t+1} \in \mathcal{A}} \hat{Q}_b \left( s_{t+1}, a_{t+1}; \theta_b^- \right) \right\} \quad (13)$$

In the following, we will stick with the naming of Waltz and Okhrin (2022) and call this algorithm *KEBDQN*. Further implementation details are deferred to the original paper.

### 4.3. Controller design for inland ASVs

In this section, we describe the state, action and reward structure of the MDP used to model inland waterways for this study. As described in Section 4.1, on every time step, $t$, the agent -our vessel- observes a state $s_t$ from the environment and chooses to perform action $a_t$ according to its policy $\pi_\theta$.

*State space* The state space $s_t = \left( s_t^{d\top} \ s_t^{e\top} \right)^\top$ is assumed to be fully observable and involves two parts: The first part

$$s^d = \left( u_t, \ v_t, \ r_t, \ \delta_t, \ u_{t-1}, \ v_{t-1}, \ r_{t-1}, \ \delta_{t-1} \right)^\top, \quad (14)$$

contains information about surge, sway and yaw rates and the rudder angle, $\delta$, at the current and previous time steps. This way the agent can perceive the changes in dynamics resulting from different environmental conditions, for example, increased sway rates due to cross-current fields, or due to the agent's actions. The second part encodes information about the surroundings of the agent:

$$s^e = \left(\tilde{y}_{e,t},\ \tilde{y}_{e,t-1},\ \chi_{e,t},\ \chi_{e,t-1},\ \frac{h_t - d}{\max(h)},\ \gamma_{rel}\right)^\top, \quad (15)$$

where $h_t$ is the current water depth below keel, and $d$ is the ship draught, thus $\frac{h_t - d}{\max(h)}$ is the remaining water under keel normalized by the maximum depth possible in the environment. The current attack angle relative to the bow is $\gamma_{rel}$, and $\tilde{y}_{e,t} = c_1 \tanh(y_{e,t})$, with $c_1$ being a tunable hyperparameter controlling the importance of the cross-track error. The above cross-track-error scaling is done to stabilize training in later stages, as its magnitude exceeds all other observations being measured.

*Action space* In this study we follow other researchers Moreira et al. (2007); Amendola et al. (2019); Zhao et al. (2019); Amendola et al. (2020) and assume constant thrust by fixing the propeller rotation rate to $4.0 s^{-1}$ i.e. the agent does not control its velocity, but its rudder angle. There are three possible actions $a_t \in \{\delta_{t-1} - 2°, \delta_{t-1}, \delta_{t-1} + 2°\}$, that either increase or decrease the rudder angle by two degrees or leave it as is. The admissible rudder range is $\delta_t \in \{-20°, -18°, \ldots, 18°, 20°\}$. The choice of a stepwise rudder change rather than choosing between fixed angles avoids generating successive rudder commands of unrealistic magnitude, for example, $\{a_t = -20°, a_{t+1} = 20°\}$, which would lead to structural damage of the rudder.

*Reward structure* The reward the environment emits acts as an immediate measurement of the quality of the action taken by the agent. To fulfill the path-following objective, minimal spatial and angular deviation from the given path is intended. Therefore, the reward system includes three parts:

$$R_t = \omega_1 R_{y_e,t} + \omega_2 R_{\chi_e,t} + R_{\text{aground},t}. \quad (16)$$

The first part rewards closeness to the desired path, while the second guides the agent towards its desired course as dictated by the underlying vector field. The terms are defined as:

$$R_{y_e,t} = \exp(-c_2 |y_{e,t}|), \quad (17)$$

and

$$R_{\chi_e,t} = \exp(-c_3 |\chi_{e,t}|) \quad (18)$$

If the water depth below the agent is less than $1.2d$, the agent will receive a negative reward defined by

$$R_{\text{aground},t} = \begin{cases} -20 & \text{if } h_t < 1.2d \\ 0 & \text{otherwise.} \end{cases} \quad (19)$$

The factor of 1.2 is used as a lower bound as the shallow-water correction terms for the hydrodynamic derivatives Kijima and Nakiri (1990); Ankudinov et al. (1990) lead to unrealistic vessel behavior below this bound. If the vessel advances to areas where the water depth falls below this threshold, the current episode is terminated. Preliminary testing concluded that values of $c_2 = 0.1$, $c_3 = 10$, $\omega_1 = 0.6$ and $\omega_2 = 0.4$ yielded a reward structure sensitive to cross-track error deviations of more than one ship width. Figure 5 shows a contour plot of the reward structure described above. The selection of weights was chosen such, that cross-track deviations are penalized more quickly than course deviations. This was done to allow the vessel to advance through curves and current fields with a non-zero drift angle while still attaining high rewards.

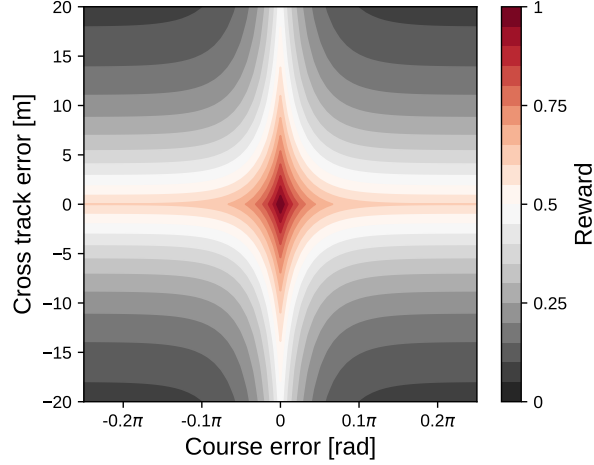

**Figure 5:** Reward contours for the path-following setup

### 4.4. Training environment generation

Since restricted waterways in general, and rivers in particular feature a wide variety of widths, lengths, riverbed profiles, water depth distributions and current velocities, a robust agent needs to be trained in an equally diverse training environment.

To generate arbitrary rivers we loosely follow the procedures in Fossen (2011, p. 255) by using an alternating sequence of straight and curved segments of equal width $w_S$, as shown in Figure 6. A given straight segment is described by the tuple $S^S := (\xi, l)$, while each curved segment is defined by the triple $S^C := (\xi, r_C, \phi)$, where $\xi$ is the starting angle of the segment against the ordinate, $l$ is the length of the straight segment, $r_C$ is the radius of the circle inscribing the curved segment, and $\phi$ is the angle by which we want the segment to curve (curvature).

A training environment is now build by chaining $n$ straight and curved segments together in an alternating fashion to form a $n$-random river

$$\text{Riv}(n) = (S_1^S, S_1^C, S_2^S, S_2^C, \ldots, S_n^S, S_n^C), \tag{20}$$

by the following rules: The first angle $\xi_1$ is initialized arbitrarily, in our study we use $\xi_1 = 0$. All successive angles are calculated by:

$$\xi_k = \xi_1 + \sum_{i=1}^{k-1} \phi_i, \tag{21}$$

for $k \in \{1, 2, \ldots, n\}$.

We additionally divide the entire $n$-random river into $p$ cross-sections $C_j = \{q_{1,j}, \ldots, q_{m,j}\}, j \in \{1, 2 \ldots, p\}$, each holding $m$ supporting points $q_{i,j} = (x_{q_{i,j}}, y_{q_{i,j}})^\top$, $i \in \{1, 2 \ldots, m\}$. The set of all supporting points forms a two-dimensional grid (see Figure 6, bottom right), which will be used for current field and water depth sampling. On straight segments, the grid is equidistant such that $d(q_{i,j}, q_{i,j+1}) = d(q_{i+1,j}, q_{i,j})$, while for curved segments, the distance between adjacent supporting points varies depending on the segment's curvature.

For every cross-section $C_j$, the water depth is sampled according to

$$h_{q_{i,j}} = -h_{\max} \exp\left\{-\epsilon \cdot d(q_{i,j}, q_j^M)^4\right\} + \eta, \tag{22}$$

with random noise $\eta \sim \mathcal{N}(0, \sigma)$, maximum water depth $h_{\max}$, and $\epsilon$, a parameter controlling the river wall steepness and fairway width. $q^M$ is the middle point of a given cross-section $j$ such that $d(q_1, q_j^M) = d(q_m, q_j^M)$.

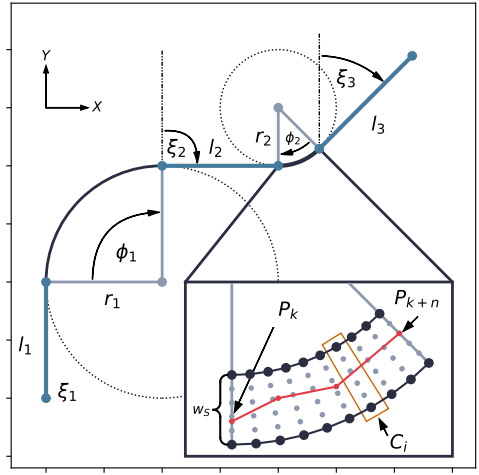

**Figure 6:** Example river generated from five segments as described in Section 4.4. The bottom right view details a curved river segment, showing the width of the segment $w_S$, a cross-section, $C_i$ (see Figure 7 for a side-view), as well as an example path comprised of four waypoints starting at $P_k$ and ending at $P_{k+n}$.

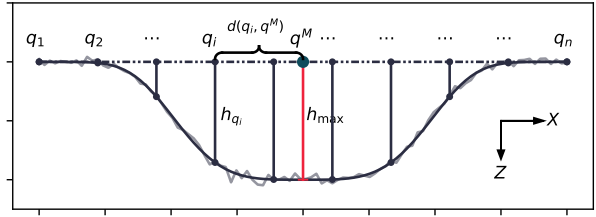

**Figure 7:** Cross-section view through a river segment. The distorted grey line resembles the depth-generation function with added noise.

To induce a current field, for a given maximum current speed $v_{max}$, and a cross-section $C_j$, we set the direction of current for all supporting points in that cross-section to be $\gamma_{q_{\cdot,j}} = \frac{2}{p}\pi j$ radians and the current speed to be $v_{q_{\cdot,j}} = f(\frac{2}{p}\pi j)v_{max}$ for all supporting points of $C_j$. The function $f : \mathbb{R} \to [-1, 1]$ can be an arbitrary continuous periodic function, this study uses the cosine. By tying the current generation process to the number of cross-sections, two rivers constructed from identical segments also share an identical current field, which is helpful in terms of reproducibility, yet, since the segments are rotated at random on each generation iteration, the likelihood of constructing alike rivers during training decreases exponentially with the number of segments.

### 4.5. Training

For training, we chose a discretization time-step of $\Delta T = 1s$ and an episode length of 2000 steps equating to roughly 33 minutes in real-time. At the beginning of each episode, a random river is generated as described in 4.4. We construct $n = 5$ straight and curved segments with angles, radii, and lengths drawn uniformly from the following sets

$$\phi \in \{\pm 60°, \pm 61°, \ldots, \pm 100°\},$$
$$r \in \{1000, \ldots, 5000\},$$
$$l \in \{400, \ldots, 2000\}.$$

The value ranges for $r$ and $l$ are chosen such that they resemble real-world river behavior and avoid the construction of unrealistically sharp turns or too short straights. During training, we sample values from each set with equal probability.

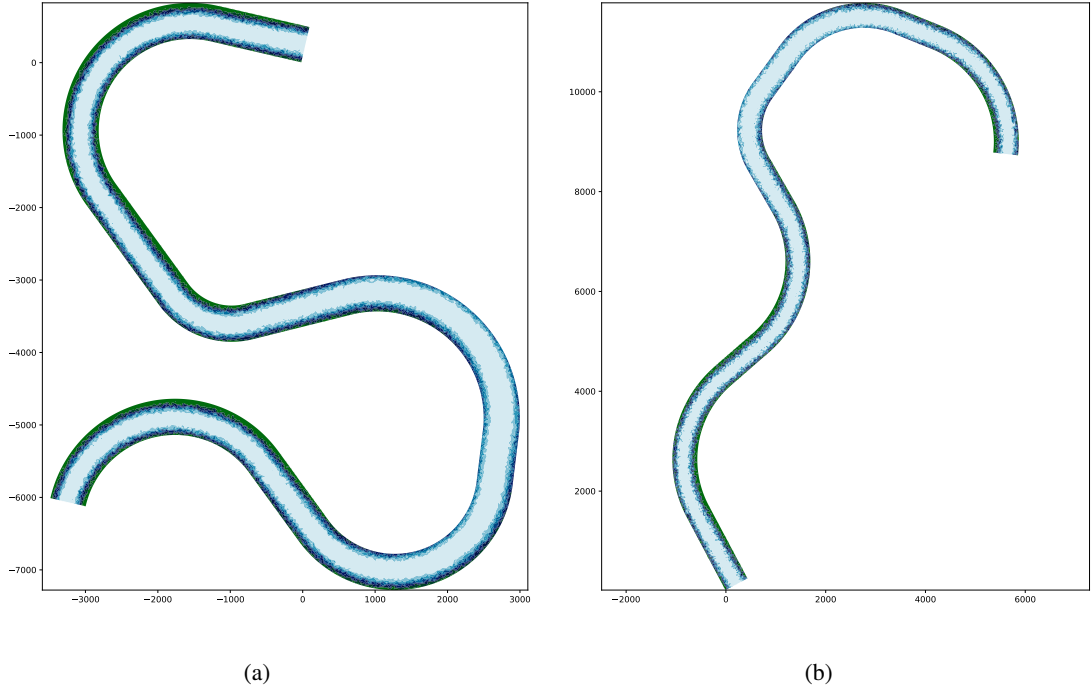

Figure 8: Example rivers used for training, generated as in 4.4

We set $w_S = 500m$, $d(q_k, q_{k+1}) = 20m$ and the maximum current velocity $v = 1.5ms^{-1}$; two example generated rivers can be found in Figure 8.

At the beginning of each episode, the agent-vessel is placed at the outset of the first segment of the constructed river with a heading equal to the current path heading plus some noise

$$\psi_0 = \chi_{P_0} + R, \quad \text{with } R \sim \mathcal{U}(-5°, 5°), \tag{23}$$

an initial speed $U_0 = 4.0ms^{-1}$, and a fixed propeller rotation rate of $4.0s^{-1}$. For this study, we use a network with one common core and 10 heads. The core network is a multilayer perceptron with a single hidden layer containing 128 neurons. The heads follow the same structure as the core, with one hidden layer, each containing 128 neurons. During training random batches of 128 transitions are sampled from a replay buffer of size $10^6$, gradient updates are performed by the Adam optimizer Kingma and Ba (2015) with a learning rate of $\alpha = 5 \times 10^{-4}$ and a discount rate of $\gamma = 0.99$. Training has been conducted for $3 \times 10^6$ steps, the implementation framework for the *KEBDQN* is the TUD_RL package Waltz and Paulig (2022) written in Python.

For comparison, we also trained a vanilla *DQN* alongside the *KEBDQN*. The DQN hyperparameter setup can be found in the appendix, while Figure 9 summarizes the training of 15 different seeds per algorithm.

## 5. PID Benchmark

In preparation for the validation of our approach, we chose a PID rudder controller design for the KVLCC2 tanker from Paramesh and Rajendran (2021) to serve as a performance benchmark. The original PID implementation is tuned to the full-size vessel, therefore the provided gains cannot be used in this paper. To find the best possible PID configuration for comparison against the DRL controller, we use a Particle Swarm Optimization (PSO) procedure with random uniform inertia weights proposed by Eberhart and Shi (2000) to tune our PID controller. The rudder angle at every time step evaluates to:

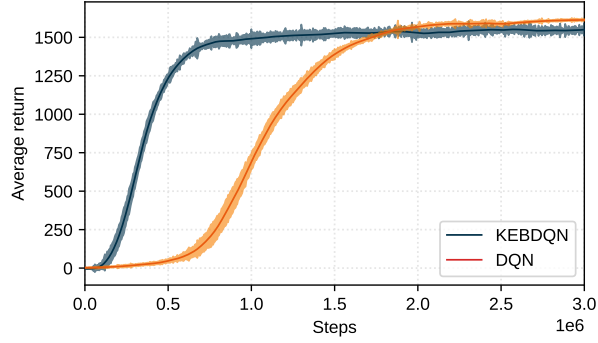

**Figure 9:** Training results running 15 independent seeds for $3 \times 10^6$ steps each. The shaded area resembles the 95% point-wise confidence intervals. The theoretical reward limit is 2000.

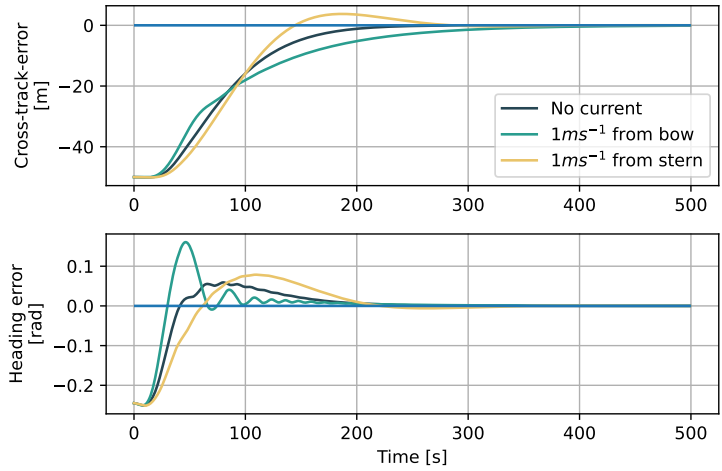

**Figure 10:** PID-response for achieving zero course error.

$$\delta_t = K_p \chi_{e,t} + K_d r_t + K_i \int_0^t \chi_{e,t} \mathrm{d}t. \tag{24}$$

Initial gains, $K_p = 2.96, K_d = 19, K_i = 0.03$, have been found via a coarse grid search. The PSO algorithm used the objective function

$$J(t) = \int_0^t \chi_{e,t}^2 \mathrm{d}t \tag{25}$$

to solve for $\mathrm{argmin}_{K_p, K_i, K_d} J(t)$ which yielded $K_p = 2.81, K_d = 64, K_i = 0.0$ as gains, thereby reducing the system to a PD controller. We also used a different objective function with an additive term for minimum overshoot, yet the result could not beat the simple procedure from above. The controller response was tested in three different scenarios. In all three, the agent is set into a straight channel with a course error of 14° and a cross-track error of 50 meters. Responses for no current, current to bow and stern can be found in Figure 10. Additionally, as with the RL agent, the maximum change in rudder angle is limited to $2° s^{-1}$ to respect the structural integrity of the ASV.

## 6. Simulation and validation

The policy found from training was simulated on several sections of the lower and middle Rhine as well as on artificial scenarios checking for reactivity under harsh environmental changes. All experiments are enrolled for the *KEBDQN*, PID and DQN approach for comparison.

### 6.1. Rhine river

The first scenario validates the performance on a near 180° degree turn on the *lower Rhine* close to Düsseldorf harbor ($51.22°\,N, 6.73°\,E$), as it features one of the tightest turns in the lower Rhine. Figure 11 shows a map containing the path to be followed, and the trajectories generated by each approach; the corresponding metrics are depicted in Figure 12.

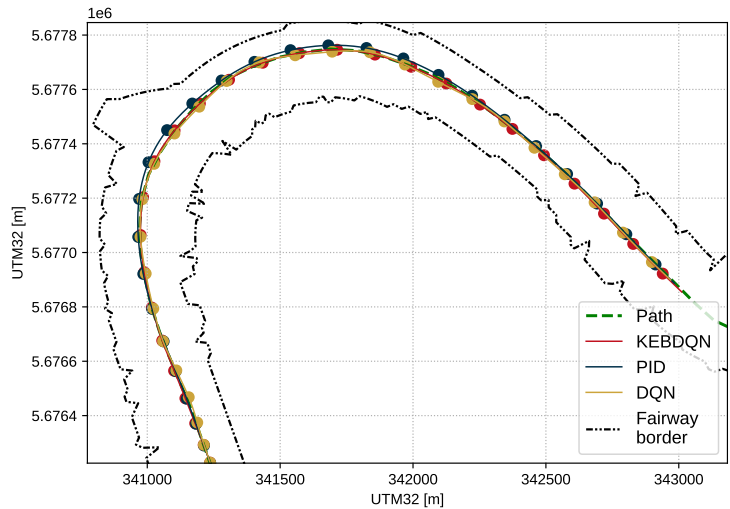

Figure 11: 2D map of the 180-degree turn around Düsseldorf harbor. The lines represent the paths taken by our ASV given the respective control approach. The dots represent equitemporally-spaced points with a distance of 30 seconds.

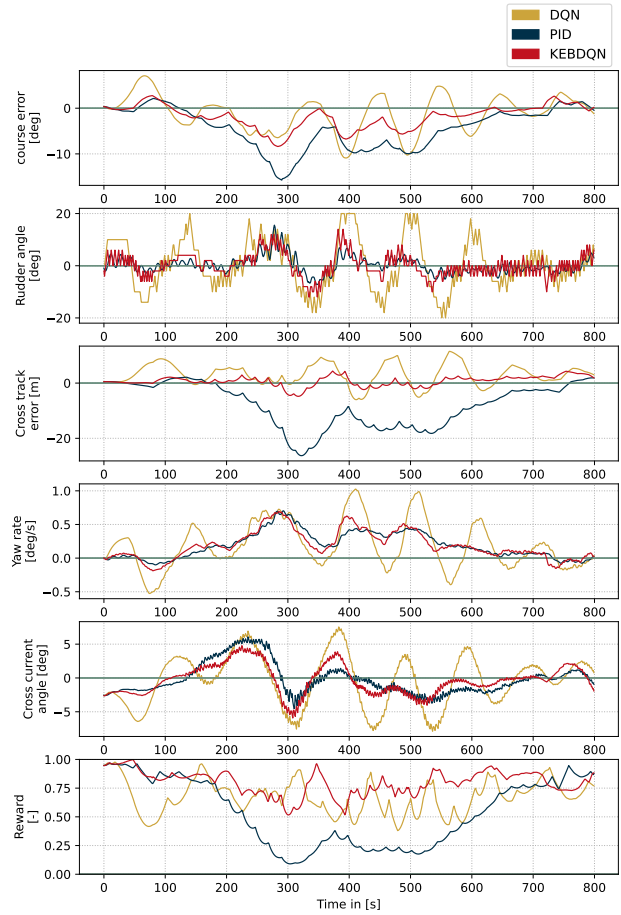

Figure 12: Time series of relevant metrics for the Düsseldorf harbor scenario. The reward plot for the PID controller is the reward it would have received if being judged by the same reward function as the RL-based controller.

The second validation scenario was selected on the *middle Rhine*. We chose a segment close to the *Lorelei* ($50.12°N, 7.73°E$) which features one of the smallest widths on the river together with a fast succession of right and left turns. The results can be seen in Figure 13 and 14. In both scenarios, the path was generated by selecting the deepest point for every cross-section through the entire river and smoothing the result using two-dimensional exponential smoothing.

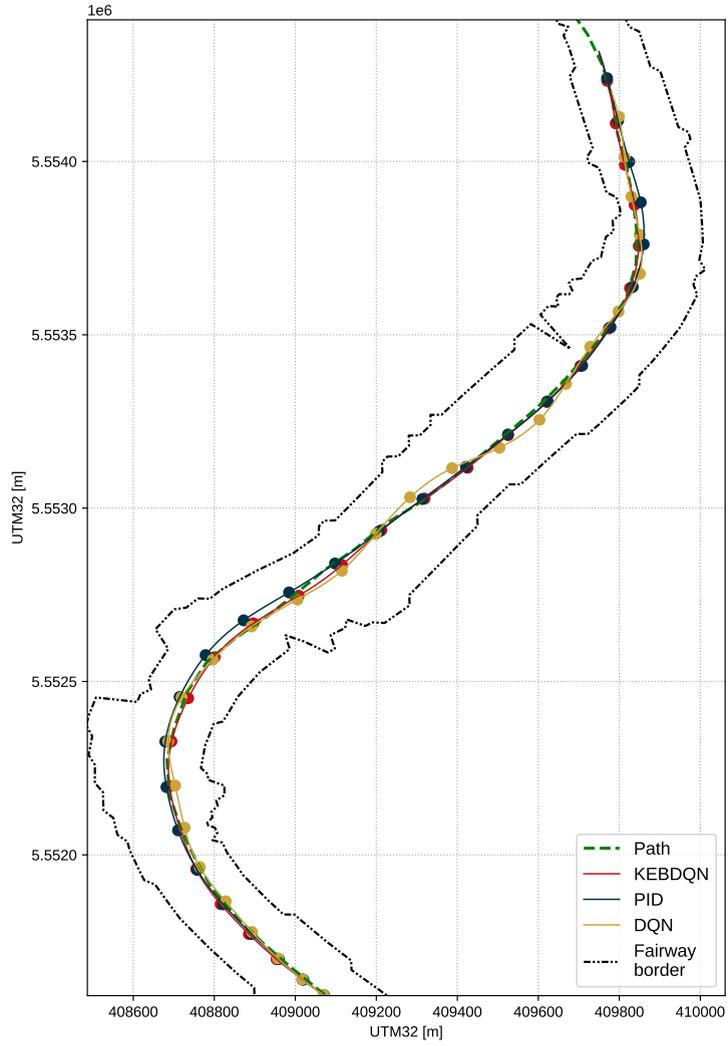

Figure 13: 2D map of the starboard turn near the Lorelei.

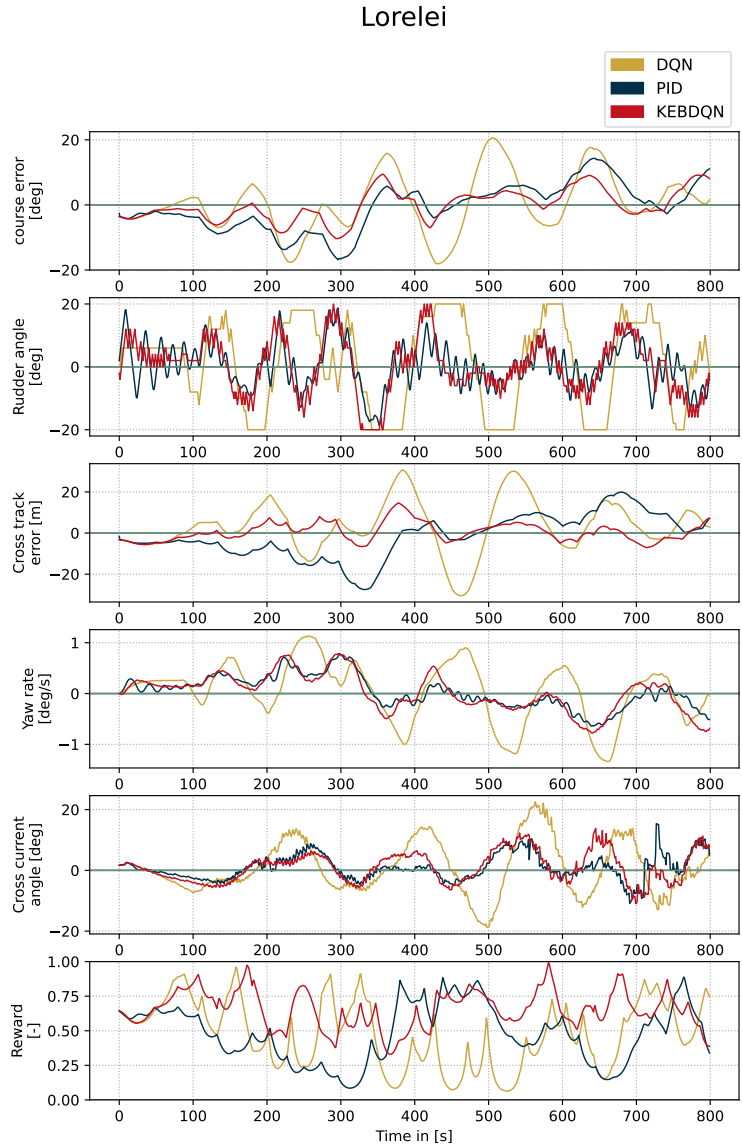

Figure 14: Time series of relevant metrics for the Lorelei scenario. The reward plot for the PID controller is the reward it would have received if being judged by the same reward function as the RL-based controller.

Analyzing the rudder commands generated for both scenarios, we observe relative similarity in magnitude and direction, indicating that the DRL agents were able to learn a similar behavior as exerted by the PID controller.

Inspecting the cross-track error and course error for both scenarios, both controllers are again found to follow akin patterns, however, the DRL controller reacts quicker, thus being able to achieve a maximum cross-track deviation of $4.36m$ compared to $26.30m$ from the PID controller for the Düsseldorf harbor scenario, and $14.67m$ and $27.47m$ respectively for the Lorelei.

One of the major drawbacks of discrete RL-based controllers is the jittering of the rudder angle as seen in the rudder commands in Figures 12 and 14, however, since the rudder steps in the RL approach are chosen such that the structural limits of the ASV are respected, the jittering is not prone to damaging the rudder of the ASV if this algorithm had been deployed in the real world. In earlier stages of research, we followed other authors Martinsen and Lekkas (2018a,b), and added a penalty for changing rudder angles too quickly, concluding that less change in rudder angle came at the cost of losing cross-track-error accuracy. Since we valued accuracy higher than slow rudder change, the penalty term was removed.

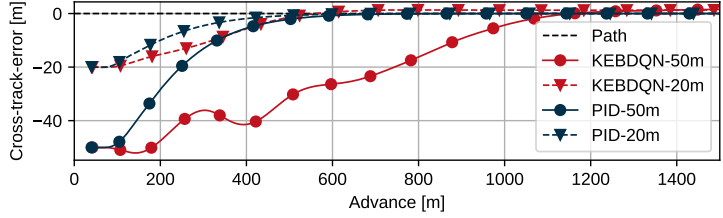

**Figure 15:** Straight-path-following experiment. Consecutive markers are each 30 seconds apart.

## 6.2. Straight paths

For maneuvers like berthing, docking, and locking or advancing through canals it seems natural to ask the vessel to follow a straight line with very high accuracy. We will test this ability for the PID and DRL controller in a straight-path scenario. We would expect the PID controller to achieve near-perfect convergence to the path, as any other result would indicate a misconfigured set-point. As with the PID calibration, the vessel will be placed in a straight canal 50 and 20 meters starboard to the path with a course error of 14° and 5.7° respectively. Propeller revolutions are fixed to $4.0s^{-1}$ and the initial velocity $U_0 = 2.0ms^{-1}$. We assume no currents and a water depth to draught ratio of roughly $h/d = 2.40$.

The results from Figure 15 confirm our initial assumption about the PID controller. In comparison to our DRL approach, the PID converges faster and more accurately, falling below one meter of cross-track error after 600$m$ of advance. The DRL rate of convergence seems to be dependent on the offset magnitude from the path. In the 20$m$ offset scenario, the *KEBDQN* performs similarly to the PID, while for the 50$m$ offset the DRL agent has noticeable difficulties returning to the path. We assume, that the agent rarely saw cross-track errors this large at the beginning of an episode during training, therefore no convergence strategy could be developed.

## 7. Robustness analysis

### 7.1. Varying revolutions

To verify the robustness of our approach, both controllers were driven up and downstream through the entire lower- and middle Rhine, each in a single episode. We did this once for a propeller revolution rate of $4.0s^{-1}$, which is also the frequency used for training, and another time using $5.0s^{-1}$ to investigate the generalization capabilities of our approach.

The cross-track-error distributions achieved are depicted in Figure 17. For the downstream scenario we find acceptable results for both controllers, whereby the DRL solution exerts significantly smaller variance, yet in the downstream scenario, is biased towards starboard, for both $4.0s^{-1}$ and $5.0s^{-1}$. Interestingly, this bias does not appear in the upstream scenarios, ruling out doubt about starboard-biased training, as we would expect to observe a bias towards port for driving upstream. For the upstream scenario, the PID controller appears to be sensitive to changes in ship velocity, with a tendency of becoming more stable for higher velocities. The inability of the PID controller to stay on course at a slower speed and high current velocities to the bow (the middle Rhine features current velocities of up to $2.4ms^{-1}$, and the lower Rhine up to $1.5ms^{-1}$), is likely due to misconfiguration of the PID controller for such environments. The fundamental problem with PIDs is that it may be impossible to find a set of gains, optimally controlling the rudder in dynamic environments featuring a wide range of external disturbances.

DRL approaches in contrast have the ability to adapt to more general cases, as they can rely on their experience acquired from the training. Although the agents did not see current velocities above $1.5ms^{-1}$, they were trained to react to those from every direction. This may lead to an additional environmental awareness, capable of achieving small cross-track errors across varying external disturbances.

### 7.2. Noisy observations

To further explore the robustness of our approach against the PID controller, we decided to compare cross-track-error performance under noisy sensor inputs. Impaired sensor measurements appear regularly in real-world applications, thus posing a valuable platform to evaluate controller behavior.

We again chose the unaltered Düsseldorf scenario on the lower Rhine as in Section 6.1 but with added Gaussian noise to the yaw-rate $\bar{r}_t = r_t + \epsilon_r, \epsilon_r \sim \mathcal{N}(0, \bar{\sigma}_r)$ and course error $\bar{\chi}_t^e = \chi_t^e + \epsilon_{\chi_e}, \epsilon_{\chi_e} \sim \mathcal{N}(0, \bar{\sigma}_{\chi_e})$. The standard deviations are calculated from the empirical distributions of yaw rate and course error obtained from driving the *KEBDQN* controller through the entire river in a single episode and had been measured to be $\bar{\sigma}_r = 0.004 \, \text{rad} \cdot s^{-1}$ and $\bar{\sigma}_{\chi_e} = 0.052°$. All other sensor inputs for the DRL approach are left unchanged.

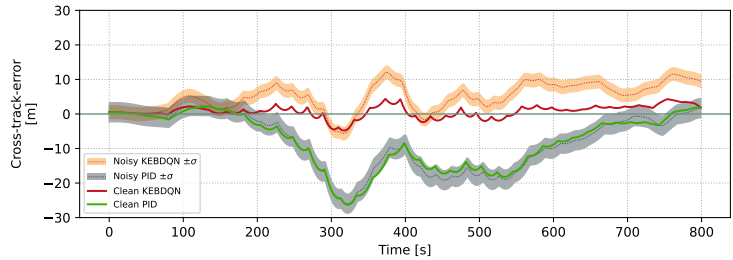

**Figure 16:** Cross-track-error for 10 different attempts per controller with sensor noise. The shaded area for the noisy runs resembles one standard deviation distance from the 10-run average. Clean runs are the same as in Figure 12 and have no noise applied to any input.

The results in Figure 16 paint an ambiguous picture. On the one hand, we still observe greater accuracy in cross-track error for the DRL controller, on the other hand, the deviation from the noise-free run is smaller for the PID controller. We also did this for several other scenarios (available upon request), all with similar outcomes. Although the variation in cross-track error for the PID controller under noisy inputs is smaller technically, the accuracy achieved by the DRL controller remains higher.

Therefore, we can conclude, that in terms of sensor noise, our PID controller attains lower deviations from its anticipated position contrary to the DRL system. However, the broader picture of robustness can be seen in favor of the DRL controller, since it not only produces smaller absolute cross-track errors -even under noisy inputs-, it also performed well under unseen propeller revolutions and very slow vessel advance rates.

## 8. Conclusion

ASV path-following on inland waterways poses several additional challenges compared to the open sea. The present study addresses these challenges by using a state-of-the-art bootstrapped DQN algorithm to develop a robust and versatile rudder controller for path-following on inland waterways. Optimal control approaches such as PID or traditional DRL algorithms such as DQN showed inferior adaptability to the highly dynamic river environment, especially for upstream scenarios with strong flow velocities to the vessel's bow. We acknowledge that those approaches can also generate rudder commands leading to accurate control of the ASV. Yet, they would require re-training or reconfiguration to adapt to the versatile dynamics of restricted waterways. Furthermore, our paper neither considers traffic nor dynamic changes in propeller revolutions, which may be oversimplified and should be addressed in future research.

## 9. Acknowledgements

The authors would like to thank the German Federal Waterways Engineering and Research Institute (BAW, Bundesanstalt für Wasserbau) for providing real-world depth and current data for the lower- and middle Rhine as well as the Center for Information Services and High-Performance Computing at TU Dresden for providing its facilities for high throughput calculations. The authors would also like to extend their gratitude to Martin Waltz and Fabian Hart for their valuable discussions and support throughout this project.

## References

Amendola, J., Miura, L.S., Costa, A.H.R., Cozman, F.G., Tannuri, E.A., 2020. Navigation in restricted channels under environmental conditions: Fast-time simulation by asynchronous deep reinforcement learning. IEEE Access 8, 149199–149213.
Amendola, J., Tannuri, E.A., Cozman, F.G., Reali Costa, A.H., 2019. Port channel navigation subjected to environmental conditions using reinforcement learning, in: International Conference on Offshore Mechanics and Arctic Engineering, American Society of Mechanical Engineers. p. V07AT06A042.

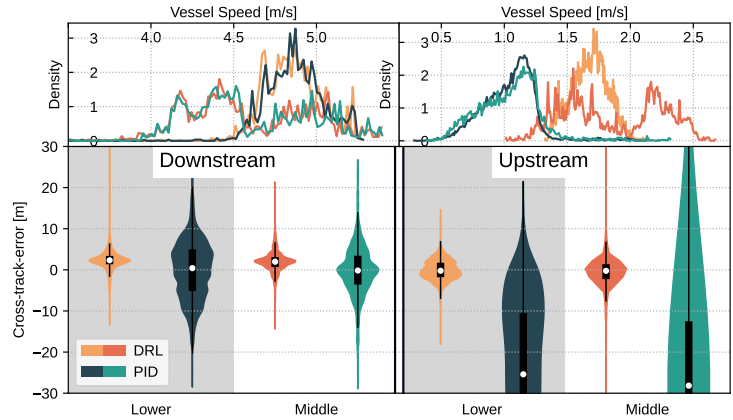

(a) Cross-track-error and speed-over-ground distributions at training revolutions of $4.0s^{-1}$. The PID controller episode in the upstream scenario on the middle Rhine did not complete the entire river due to running aground.

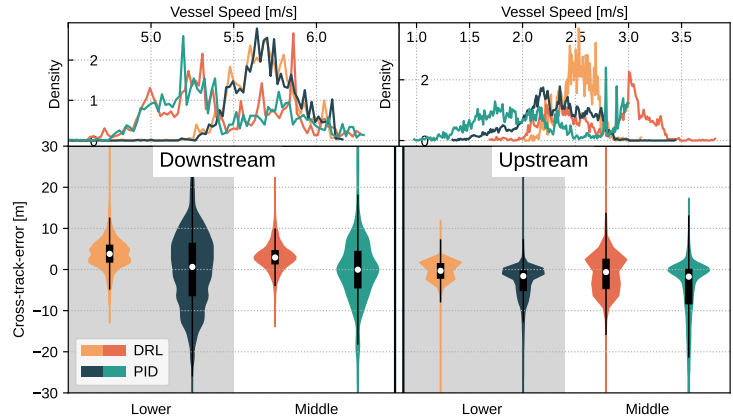

(b) Cross-track-error and speed-over-ground distributions for unseen revolutions of $5.0s^{-1}$

**Figure 17:** Cross-track-error and speed-over-ground distributions for the lower- and middle Rhine for the DLR and PID controller at $4.0s^{-1}$ (a) and $5.0s^{-1}$ (b) revolutions. The white dots represent the median of the distribution and the black bars correspond to the interquartile range. The plots at the top, represent the distributions of vessel speeds for each attempt. Kernel densities were estimated using gaussian kernels with bandwidths calculated after Botev et al. (2010).

Amin, O.M., Hasegawa, K., 2010. Generalised mathematical model for ship manoeuvrability considering shallow water effect, in: Conference Proc, of Japan Society of Naval Architects and Ocean Engineers. pp. 531–534.

Ankudinov, V., Miller, E., Jakobsen, B., Daggett, L., 1990. Manoeuvring performance of tug/barge assemblies in restricted waterways. Proceedings MARSIM & ICMS 90, 515–525.

Bellman, R., 1957. Dynamic programming, princeton univ. Press Princeton, New Jersey .

BIS Research, 2018. Global autonomous ship and ocean surface robot market: Analysis and forecast, 2018-2028. URL: https://www.whatech.com/markets-research/it/archive/521945-global-ocean-surface-robot-market-anticipated-to-reach-2-90-billion-by-2028-at-a-cagr-of-16-8-and-global-autonomous-ship-market-expected-to-generate-a-cumulative-revenue-of-3-48-billion-by-2035-bis-research-report.

Botev, Z., Grotowski, J., Kroese, D., 2010. Kernel density estimation via diffusion. The Annals of Statistics , 2916–2957.

Breivik, M., Fossen, T.I., 2004. Path following of straight lines and circles for marine surface vessels. IFAC Proceedings Volumes 37, 65–70.

Directorate-General for Mobility and Transport, 2023. Inland waterways. URL: https://transport.ec.europa.eu/transport-modes/inland-waterways_en.

Eberhart, R.C., Shi, Y., 2000. Comparing inertia weights and constriction factors in particle swarm optimization, in: Proceedings of the 2000 congress on evolutionary computation. CEC00 (Cat. No. 00TH8512), IEEE. pp. 84–88.

Fossen, T.I., 2011. Handbook of marine craft hydrodynamics and motion control. John Wiley & Sons.

Fossen, T.I., Breivik, M., Skjetne, R., 2003. Line-of-sight path following of underactuated marine craft. IFAC proceedings volumes 36, 211–216.

Fossen, T.I., Lekkas, A.M., 2017. Direct and indirect adaptive integral line-of-sight path-following controllers for marine craft exposed to ocean currents. International journal of adaptive control and signal processing 31, 445–463.

Kijima, K., Nakiri, Y., 1990. Prediction method of ship maneuverability in deep and shallow waters. Proceedings of MARSIM and ICSM 90 .

Kingma, D.P., Ba, J., 2015. Adam: A method for stochastic optimization, in: Bengio, Y., LeCun, Y. (Eds.), 3rd International Conference on Learning Representations, ICLR 2015, San Diego, CA, USA, May 7-9, 2015, Conference Track Proceedings. URL: http://arxiv.org/abs/1412.6980.

Liu, Y., Bu, R., Gao, X., 2018. Ship trajectory tracking control system design based on sliding mode control algorithm. Polish Maritime Research , 26–34.

Martinsen, A.B., Lekkas, A.M., 2018a. Curved path following with deep reinforcement learning: Results from three vessel models, in: OCEANS 2018 MTS/IEEE Charleston, IEEE. pp. 1–8.

Martinsen, A.B., Lekkas, A.M., 2018b. Straight-path following for underactuated marine vessels using deep reinforcement learning. IFAC-PapersOnLine 51, 329–334.

Martinsen, A.B., Lekkas, A.M., Gros, S., Glomsrud, J.A., Pedersen, T.A., 2020. Reinforcement learning-based tracking control of usvs in varying operational conditions. Frontiers in Robotics and AI 7, 32.

Mnih, V., Kavukcuoglu, K., Silver, D., Rusu, A.A., Veness, J., Bellemare, M.G., Graves, A., Riedmiller, M., Fidjeland, A.K., Ostrovski, G., et al., 2015. Human-level control through deep reinforcement learning. nature 518, 529–533.

Moreira, L., Fossen, T.I., Soares, C.G., 2007. Path following control system for a tanker ship model. Ocean Engineering 34, 2074–2085.

Nelson, D.R., Barber, D.B., McLain, T.W., Beard, R.W., 2007. Vector field path following for miniature air vehicles. IEEE Transactions on Robotics 23, 519–529.

Oh, S.R., Sun, J., 2010. Path following of underactuated marine surface vessels using line-of-sight based model predictive control. Ocean Engineering 37, 289–295.

Osband, I., Blundell, C., Pritzel, A., Van Roy, B., 2016. Deep exploration via bootstrapped dqn. Advances in neural information processing systems 29.

Paramesh, S., Rajendran, S., 2021. A unified seakeeping and manoeuvring model with a pid controller for path following of a kvlcc2 tanker in regular waves. Applied Ocean Research 116, 102860.

Paulig, N., 2022. Mmg standard model for ship maneuvering. https://github.com/nikpau/mmgdynamics.

Perera, L.P., Ferrari, V., Santos, F.P., Hinostroza, M.A., Soares, C.G., 2014. Experimental evaluations on ship autonomous navigation and collision avoidance by intelligent guidance. IEEE Journal of Oceanic Engineering 40, 374–387.

Puterman, M.L., 1994. Markov decision processes: discrete stochastic dynamic programming. John Wiley & Sons Inc.

Sandeepkumar, R., Rajendran, S., Mohan, R., Pascoal, A., 2022. A unified ship manoeuvring model with a nonlinear model predictive controller for path following in regular waves. Ocean Engineering 243, 110165.

Shen, H., Guo, C., 2016. Path-following control of underactuated ships using actor-critic reinforcement learning with mlp neural networks, in: 2016 Sixth International Conference on Information Science and Technology (ICIST), IEEE. pp. 317–321.

Sutton, R.S., Barto, A.G., 2018. Reinforcement learning: An introduction. MIT press.

Taimuri, G., Matusiak, J., Mikkola, T., Kujala, P., Hirdaris, S., 2020. A 6-dof maneuvering model for the rapid estimation of hydrodynamic actions in deep and shallow waters. Ocean Engineering 218, 108103.

Thrun, S., Schwartz, A., 1993. Issues in using function approximation for reinforcement learning, in: Proceedings of the 1993 Connectionist Models Summer School Hillsdale, NJ. Lawrence Erlbaum, pp. 1–9.

Van Hasselt, H., 2010. Double q-learning. Advances in neural information processing systems 23.

Van Hasselt, H., Guez, A., Silver, D., 2016. Deep reinforcement learning with double q-learning, in: Proceedings of the AAAI conference on artificial intelligence.

Waltz, M., Okhrin, O., 2022. Two-sample testing in reinforcement learning. arXiv preprint arXiv:2201.08078 .

Waltz, M., Paulig, N., 2022. Rl dresden algorithm suite. https://github.com/MarWaltz/TUD_RL.

Watkins, C.J., Dayan, P., 1992. Q-learning. Machine learning 8, 279–292.

Woo, J., Yu, C., Kim, N., 2019. Deep reinforcement learning-based controller for path following of an unmanned surface vehicle. Ocean Engineering 183, 155–166.

Xia, G., Liu, J., Wu, H., 2013. Neural network based nonlinear model predictive control for ship path following, in: 2013 Ninth International Conference on Natural Computation (ICNC), IEEE. pp. 210–215.

Yasukawa, H., Yoshimura, Y., 2015. Introduction of mmg standard method for ship maneuvering predictions. Journal of Marine Science and Technology 20, 37–52.

Zhang, X., Yang, G., Zhang, Q., Zhang, G., Zhang, Y., 2017. Improved concise backstepping control of course keeping for ships using nonlinear feedback technique. The Journal of Navigation 70, 1401–1414.

Zhao, L., Roh, M.I., Lee, S.J., 2019. Control method for path following and collision avoidance of autonomous ship based on deep reinforcement learning. Journal of Marine Science and Technology 27, 1.

| Parameter | Value |
| --- | --- |
| Number of hidden layers | 2 |
| Number of neurons per layer | [256, 128] |
| Batch size | 128 |
| Discount factor ($\gamma$) | 0.99 |
| Loss function | MSE |
| Replay buffer size | $1.0 \times 10^6$ |
| Optimizer | Adam |
| Target network update frequency | 1000 steps |
| Initial exploration rate ($\epsilon_{initial}$) | 1.0 |
| Final exploration rate ($\epsilon_{final}$) | 0.01 |
| Exploration decay time | $1.0 \times 10^6$ steps |

**Table 2**
List of hyperparameters used to train the DQN.

## A. DQN hyperparameters